\documentclass[utf8]{FrontiersinHarvard} 
\usepackage{url,hyperref,lineno,microtype}
\usepackage[onehalfspacing]{setspace}
\usepackage{siunitx}
\usepackage{booktabs}
\usepackage{capt-of}

\def\keyFont{\fontsize{8}{11}\helveticabold }
\def\firstAuthorLast{Zhang {et~al.}} 
\def\Authors{Han Zhang\,$^{1,*}$, Lalithkumar Seenivasan\,$^{1}$, Jose L. Porras\,$^{1,2}$, Roger D. Soberanis-Mukul\,$^{1}$, Hao Ding\,$^{1}$, Hongchao Shu\,$^{1}$, Benjamin D. Killeen\,$^{1}$, Ankita Ghosh\,$^{1}$, Lonny Yarmus\,$^{2}$, Jeffrey K. Jopling\,$^{2}$, Masaru Ishii\,$^{2}$, Angela C. Argento\,$^{2}$ and Mathias Unberath\,$^{1}$}

\def\Address{$^{1}$Johns Hopkins University, Baltimore, 21218, MD, USA \\
$^{2}$Johns Hopkins Medical Institutions, Baltimore, 21287, MD, USA}
\def\corrAuthor{Han Zhang}
\def\corrEmail{hzhan206@jh.edu}

\begin{document}
\onecolumn
\firstpage{1}

\title[Egocentric view synthesis in ORs]{Egosurg: Arbitrary view synthesis for egocentric replay of operating room workflows from ambient cameras} 

\author[\firstAuthorLast ]{\Authors} 
\address{} 
\correspondance{} 

\extraAuth{}%

\maketitle

\begin{abstract}

Observing surgical practice has historically relied on fixed vantage points or recollections, leaving the egocentric perspectives that shape clinical decisions undocumented. Ambient fixed cameras capture the operating room (OR) at room scale but cannot recover what any individual team member actually saw. We present EgoSurg, a framework that reconstructs dynamic OR scenes from sparse wall-mounted stereo video and renders arbitrary, role-specific egocentric views without instrumenting personnel or interfering with clinical workflow. EgoSurg initializes a per-timestamp 3D Gaussian Splatting representation from scale-aware stereo depth and refines it with an image-conditioned diffusion model that corrects auxiliary rendered views, mitigating artifacts caused by limited camera coverage, crowding, and occlusion. We evaluated the framework on four real robotic pulmonology procedures and two simulated full-workflow sessions across two hospital sites. Near-field reconstruction fidelity was consistent (PSNR 26.8~dB, SSIM .895) across five workflow phases and both sites, and synthesized egocentric views reached a PSNR of 17.8~dB and an SSIM of .766 against paired hand-held point-of-view recordings. We further demonstrate three case studies for intended use: adjudicating a simulated sterile field violation, replaying a procedure from role-specific viewpoints, and testing a counterfactual change in personnel position. These results indicate that existing ambient camera infrastructure can be turned into a navigable 3D record of surgical work, supporting retrospective review of safety events, training, and workflow analysis from every angle.

\tiny
 \keyFont{ \section{Keywords:} novel view synthesis, 3D Gaussian splatting, diffusion models, surgical data science, operating room, ambient intelligence} 
\end{abstract}

\section{Introduction}

The very term ``operating theater'' emerged from an era of tiered amphitheaters, when learning, observing, and improving surgical practice required being \emph{there} either on the surgical team or in the gallery. Beyond those vantage points, decisions, sightlines, micro-interactions, and many findings were lost to the memory and anecdote of those privileged to witness the moment. Hospitals have since adopted structured documentation and routines, including operative notes, intraoperative nursing records, anesthesia logs, and postoperative debriefs, to capture and analyze events for clinical training and optimization. Yet this approach is hard to scale and adds burden on clinicians. With written summaries and fading memory as the primary record, fine-grained detail and context -- who stood where, what they could see, how long consensus took, which paths were blocked, and when sterile or safety envelopes were at risk -- are lost to time. 

More recently, fixed-camera recording has partly alleviated these limitations. Although easy to deploy, fixed cameras may fail to capture crucial interactions when surgical orientation changes or when crowding and occlusion obscure the scene. Multi-camera setups broaden coverage yet still restrict observers to pre-defined views, forcing them to mentally model missing vantage points and limiting both observational learning and retrospective analysis \citep{brennan_intraoperative_2023,cheikh_youssef_evolution_2023,perez_privacy-preserving_2025}. Wearable cameras promise egocentric viewpoints, but introduce sterility risks and integration challenges, and constrain perspective to previously walked motion paths \citep{ozsoy_egoexor_2025,soberanis-mukul_interpretable_2025}. Dense instrumentation such as markers or active depth sensors is impractical in the tightly constrained, dynamic operating room (OR) \citep{zhang_straighttrack_2024}.

These barriers highlight a central unmet need: the ability to retrospectively reconstruct what OR personnel could have seen, from any vantage point, without altering clinical workflow. Generating arbitrary views on demand -- for example, an egocentric view of any role in the room at any moment in the timeline -- answers a question that fixed views cannot: what was actually visible to the decision maker, and what was not? That shift enables three classes of downstream application. In \textit{training}, learners can replay a decisive instant from any team member's line of sight, replacing speculation with concrete visual evidence \citep{etherington_interprofessional_2019}, and can immerse themselves in the scene from the position of different roles \citep{killeen_stand_2024}. In \textit{efficiency}, teams can perform spatially immersive retrospective analysis for workflow optimization \citep{killeen_pelphix_2023} and ask counterfactual questions, such as whether shifting a monitor during induction would have improved line of sight and shortened handoffs. In \textit{safety}, near misses -- collisions, drape encroachments, sterile field violations -- become inspectable from the viewpoint that mattered rather than remaining anecdotal.

Recent advances in computer vision demonstrate that high-fidelity novel view synthesis is now feasible from sparse, fixed camera inputs. Neural rendering methods, including neural radiance field (NeRF) variants and 3D Gaussian Splatting (3DGS), have reduced input-view requirements and accelerated rendering while preserving spatial coherence \citep{li_neuralangelo_2023,gerats_nerf-or_2025,mildenhall_nerf_2020,masuda_os-nerf_2024,kerbl_3d_2023,shu_seamless_2025}. In parallel, diffusion models have emerged as strong generative priors capable of synthesizing photorealistic images and, critically, generating plausible auxiliary views from limited inputs \citep{zhong_taming_2025,wu_difix3d_2025}. Together, these advances now open the possibility of addressing the distinctive challenges of the OR, where dynamic motion, specialized instruments, and crowded team interactions routinely disrupt conventional pipelines.

\begin{figure}[t]
  \centering
  \includegraphics[width=\textwidth]{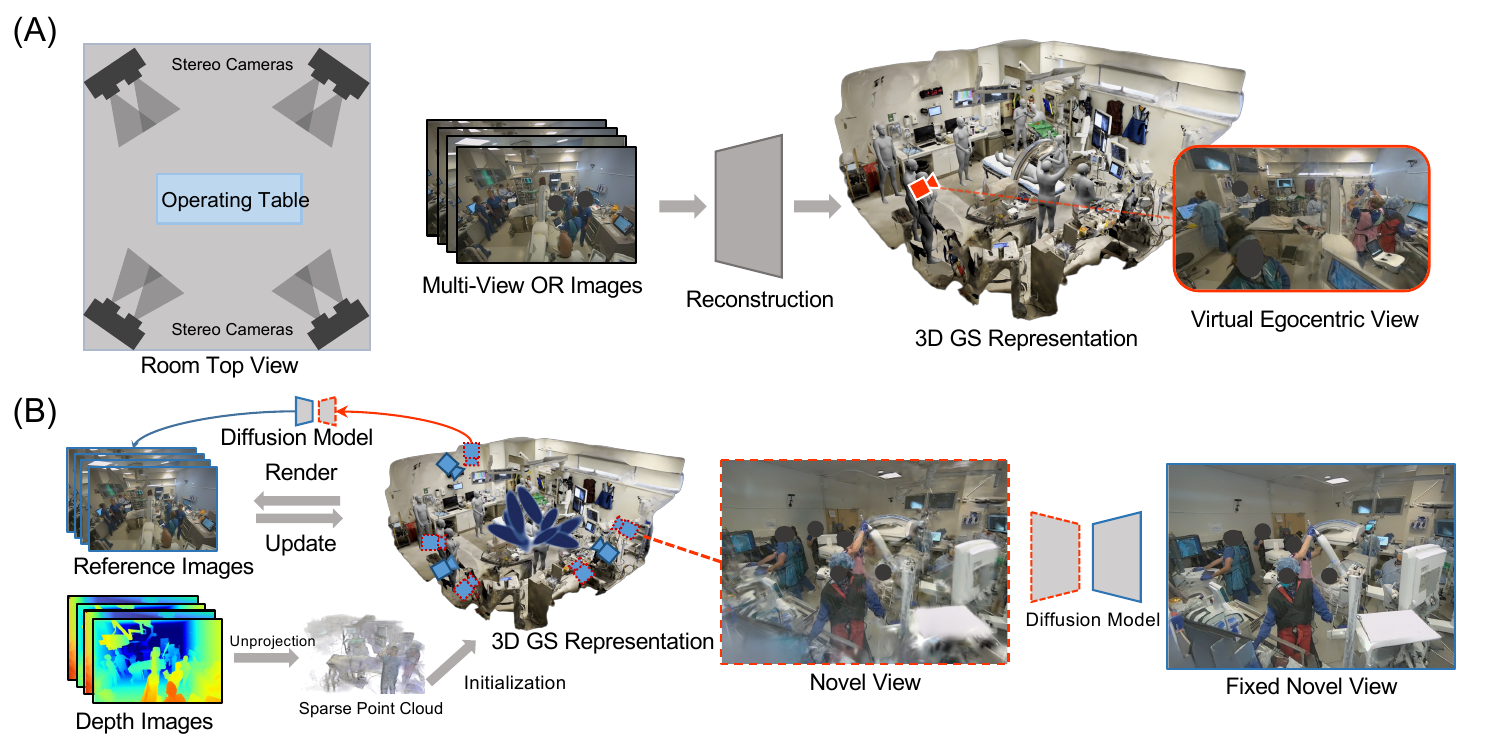}
  \caption{\textbf{(A)} Overview of the EgoSurg pipeline: Videos from sparse wall-mounted cameras are integrated into a 3DGS representation, from which virtual egocentric views can be synthesized. \textbf{(B)} Reconstruction process: Stereo depth estimates and camera calibration are used to initialize a sparse point cloud that seeds the 3DGS representation, which is then optimized against the reference images. A diffusion model generates auxiliary views to mitigate occlusions and improve cross-view consistency.}
  \label{fig:pipeline}
\end{figure}

We introduce EgoSurg, a framework that reconstructs dynamic OR scenes from synchronized wall-mounted cameras and synthesizes arbitrary views from user-specified virtual camera poses (Figure~\ref{fig:pipeline}A). At each timestamp, EgoSurg initializes a 3DGS representation from stereo depth estimates obtained from the sparse camera inputs. The 3DGS representation is then refined using auxiliary rendered views corrected by an image-conditioned diffusion model. Given a specified virtual camera pose, EgoSurg can synthesize a role-specific view that approximates the physical vantage point of an OR staff member, without requiring wearable cameras. We evaluate EgoSurg using more than five hours of recordings from four real robotic pulmonology procedures and two simulated full-workflow sessions across two sites. EgoSurg maintained comparable near-field reconstruction fidelity across workflow phases, achieving an overall PSNR of 26.8~dB and an SSIM of .895. Against paired hand-held point-of-view recordings, EgoSurg achieved a PSNR of 17.8~dB and an SSIM of .766. Finally, we demonstrate EgoSurg in three case studies: reviewing a simulated critical safety event, replaying a robotic pulmonology case, and analyzing counterfactual repositioning. By transforming routine OR video into a navigable 3D record, EgoSurg enables clinicians and trainees to visualize, experience, and analyze critical moments from every angle through perspective-agnostic replays. In essence, it allows us to ask, and answer, questions such as: ``What was happening behind the surgeon's left shoulder?''

\section{Methods}

EgoSurg reconstructs a unified OR scene representation at each timestamp from multiple wall-mounted cameras that stream continuously during a case. The framework integrates multi-camera capture, stereo depth estimation, and per-timestamp scene reconstruction (Figure~\ref{fig:pipeline}B). At each timestamp, the calibrated camera setup captures images of the OR from multiple viewpoints; FoundationStereo \citep{wen2025stereo} produces scale-aware depth maps for each stereo pair, which are unprojected into 3D space to form a sparse point cloud describing the full scene. This point cloud initializes a dense reconstruction via 3DGS, refined with an image-conditioned diffusion model that generates auxiliary views. The resulting representation supports photorealistic synthesis from arbitrary viewpoints, both egocentric and exocentric.

\subsection{Ambient Camera Infrastructure}

The ambient sensing infrastructure consists of four wall-mounted stereo RGB cameras (Stereolabs ZED X) paired with two NVIDIA Jetson Orin NX units for synchronized recording. Each stereo camera incorporates two RGB sensors with a fixed baseline, enabling metric depth estimation via disparity and triangulation. To balance unobtrusiveness and visibility, cameras were mounted above head height on walls opposite the operating table. At each timestamp the system records four stereo pairs (eight images in total), providing a multi-view representation of the scene.

Accurate extrinsic calibration and temporal synchronization are essential. Inter-camera synchronization was achieved through hardware-based Precision Time Protocol (PTP). Intrinsic parameters were adopted from factory calibration, while extrinsic parameters were estimated using a custom calibration wand consisting of three fixed spheres. By capturing wand trajectories throughout the room, cross-camera correspondences were identified. Extrinsics were first estimated using a linear relative pose solution, then refined through bundle adjustment. Global scale was recovered from the known inter-sphere distances.

\subsection{3D Gaussian Splatting Reconstruction}

3DGS offers an explicit 3D representation with high rendering efficiency and scalability, making it well suited to dynamic and cluttered OR scenes. Unlike implicit NeRF-based methods, 3DGS accommodates both static and deformable content while supporting real-time visualization. At each timestamp, synchronized stereo pairs are processed into sparse depth maps that are fused into a unified point cloud. Independent 3DGS representations are then trained per timestamp to capture the temporal evolution of the OR environment. Training proceeds in two stages: (i) scene initialization from the sparse point map, and (ii) iterative 3DGS optimization augmented with diffusion-based view refinement.

\subsubsection{Scene initialization}

Initialization is critical for stable optimization. We employ FoundationStereo \citep{wen2025stereo} with calibrated intrinsics and extrinsics to generate scale-aware depth maps from stereo pairs. These depth maps are unprojected into 3D space and merged into a shared coordinate frame. To balance geometric fidelity and efficiency, voxel downsampling and statistical outlier removal are applied. The resulting filtered point cloud serves as the initialization for 3DGS training.

\subsubsection{Scene training}

After initialization, Gaussian parameters (position, covariance, opacity, and color) are iteratively optimized to minimize the discrepancy between rendered and observed images. The training objective combines an $\ell_1$ reconstruction term, a structural similarity component (D-SSIM) \citep{kerbl_3d_2023}, and an opacity regularization term that encourages stable and well-conditioned transparency values:

\begin{equation}
L = (1 - \alpha) \, L_{1} + \alpha \, L_{\text{D-SSIM}} \; + \; \lambda_{\text{opacity}} \cdot \sigma\!\big(\mathbf{o}\big),
\end{equation}
where $\alpha$ balances pixel-wise accuracy and structural similarity, $\lambda_{\text{opacity}}$ controls the strength of regularization, $\sigma(\cdot)$ denotes the sigmoid activation, and $\mathbf{o}$ represents the set of Gaussian opacities.
 
To mitigate overfitting caused by limited viewpoints, densification, splitting, and pruning operations are disabled, and randomized backgrounds are introduced during training. This restricts optimization to Gaussian points derived from the initial point map. Training begins with a warm-up stage that uses only the sparse input views, after which optimization proceeds using both the original views and the diffusion-augmented views described in the following section.

\subsubsection{Diffusion-augmented view refinement}

To further enhance geometric consistency and mitigate artifacts arising from sparse viewpoints, we introduce diffusion-refined views during training. Around each camera, we sample $N$ poses within a hemispherical radius $R$ and render images using the current 3DGS representation. These intermediate rendered views often exhibit artifacts due to incomplete coverage and limited observation. Following Difix3D~\citep{wu_difix3d_2025}, we employ an image-conditioned diffusion model that refines each rendered view by jointly conditioning on the rendered auxiliary view and its corresponding reference image. The model is based on a single-step diffusion process derived from SD-Turbo~\citep{sauer_adversarial_2023}. In our experiments, we directly use the publicly released Difix3D weights~\citep{wu_difix3d_2025}, without additional fine-tuning. After refinement, the corrected views are incorporated back into the training set, enabling 3DGS to learn from both real and diffusion-refined views, thereby improving robustness and reducing reconstruction artifacts.

\section{Experiments}  

\subsection{Datasets}

\begin{figure}[t]
\centering
\begin{minipage}{\textwidth}
\centering
\captionof{table}{Summary of datasets used for evaluation.}
\label{tab:datasets}
\vspace{4pt}
\begin{tabular}{lllcc}
\toprule
\textbf{Dataset} & \textbf{Site} & \textbf{Procedure} & \textbf{Trials} & \textbf{Sampling} \\
\midrule
PULM Workflow     & A & Robotic pulmonology   & 4 & 25 timestamps and 5 clips per trial \\
HALO Workflow     & B & Role-play OR workflow & 2 & 25 timestamps and 5 clips per trial \\
Egocentric        & A & Routine OR activities & 1 & 10 timestamps \\
\bottomrule
\end{tabular}
\end{minipage}
\vspace{0.8em}
\includegraphics[width=0.7\textwidth]{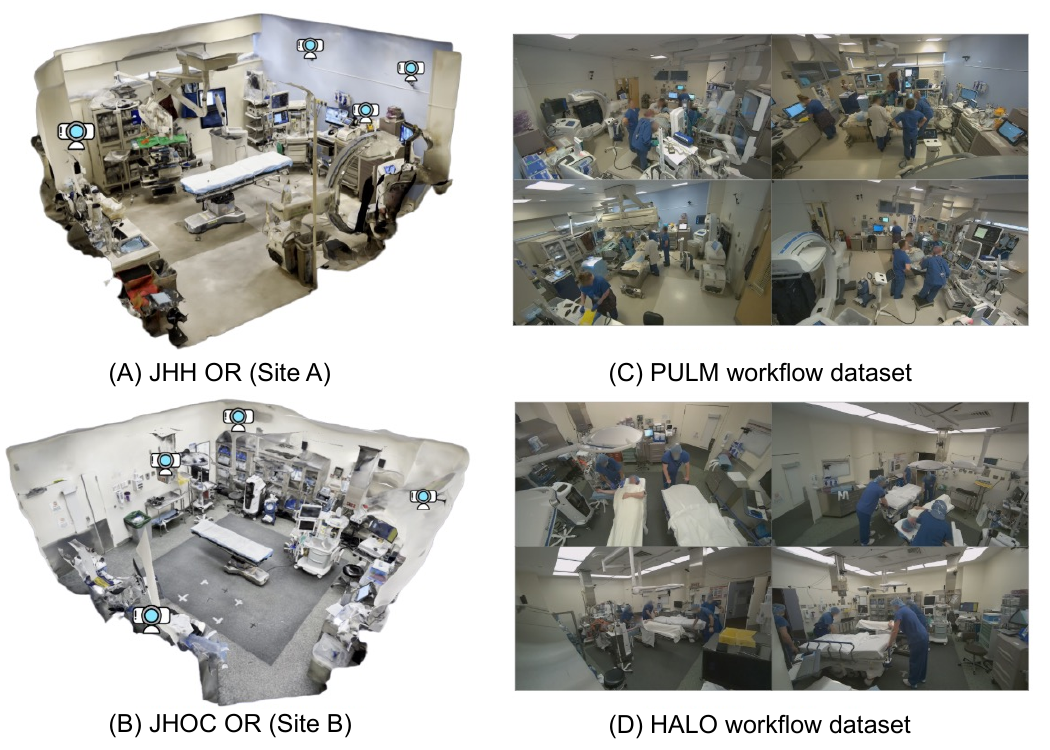}
\caption{Overview of datasets and acquisition setup.
\textbf{(A)} Johns Hopkins Hospital OR.
\textbf{(B)} Johns Hopkins Outpatient Center OR.
\textbf{(C)} Real robotic pulmonology procedure.
\textbf{(D)} Simulated role-play procedure.}
\label{fig:dataset}
\end{figure}

We evaluated EgoSurg using real surgical recordings and controlled experiments collected across two hospital sites (Figure~\ref{fig:dataset}; Table~\ref{tab:datasets}). The evaluation datasets comprised four real robotic pulmonology procedures, two simulated full-workflow procedures, and a controlled egocentric dataset with paired point-of-view (PoV) recordings. The PULM Workflow dataset includes more than five hours of recordings from four real-patient robotic pulmonology procedures performed at Johns Hopkins Hospital (Site A; Figure~\ref{fig:dataset}C), from which 25 timestamps and 5 short clips per trial were sampled for reconstruction and evaluation. These procedures involved lung biopsy using the Ion robotic platform under C-arm fluoroscopy. The combination of a crowded clinical team, mobile imaging equipment, and a surgical robot produced substantial occlusion, motion, and changes in room configuration across workflow phases. To evaluate transfer across environments, we additionally recorded two simulated full-length surgical workflows at Johns Hopkins Outpatient Center (Site B; Figure~\ref{fig:dataset}D), which differed from Site A in room geometry, equipment configuration, and workflow organization. The same sampling protocol was applied, yielding 150 reconstructed timestamps in total across both sites.

To provide ground truth for egocentric-view evaluation, we collected a controlled dataset at Site A in which a staff member carried a hand-held PoV camera while performing routine OR activities. The PoV recordings were acquired concurrently with the ambient camera streams and served as reference images for quantitative evaluation of synthesized egocentric views. All datasets were recorded using four wall-mounted stereo RGB cameras (Stereolabs ZED X). The camera systems were calibrated after installation using the procedure described in the Methods section, yielding mean reprojection errors of \SI{0.44}{px} at Site A and \SI{0.55}{px} at Site B. All data collection was conducted under Institutional Review Board approval and in accordance with applicable ethical and regulatory requirements.

\subsection{Baselines and Ablation}
We compared EgoSurg against two baseline reconstruction methods and one ablated variant:

\begin{enumerate}
    \item \textbf{Depth reprojection}: Depth maps were unprojected using the calibrated camera intrinsics and extrinsics, and fused into a unified point cloud. Novel views were generated by directly projecting the fused point cloud into the target camera.    
    \item \textbf{Image-only 3DGS}: A 3DGS representation with randomly initialized Gaussians was optimized using only the captured multi-view images, without stereo-depth initialization or diffusion-augmented supervision.
    \item \textbf{EgoSurg without diffusion}: The proposed depth-initialized 3DGS representation was optimized using only the captured multi-view images, without diffusion-refined auxiliary views.
\end{enumerate}

\subsection{Implementation Details}

All experiments were conducted on a Linux workstation equipped with an Intel Core i9-12900K CPU and an NVIDIA GeForce RTX 4090 GPU, running Ubuntu 22.04 LTS with PyTorch 2.7.1. For diffusion-augmented view generation, we sampled $N=15$ viewpoints within a radius of $R=0.4$~m around each camera. 3DGS training used 0-th order spherical harmonics and proceeded in two stages: 500 warm-up steps followed by 1500 refinement steps with augmented views. Optimization employed the Adam optimizer with a learning rate of $1.6 \times 10^{-4}$ and random background sampling. The SSIM loss weight was set to $\alpha=0.2$, and the opacity regularization weight was fixed at $0.2$.

\subsection{Evaluation Protocols}

We evaluated EgoSurg under three quantitative protocols: (i) near-field reconstruction fidelity across workflow conditions, (ii) egocentric-view fidelity against paired PoV recordings, and (iii) frame-to-frame stability over time. Where ground-truth images were available, we measured peak signal-to-noise ratio (PSNR), structural similarity index (SSIM), and learned perceptual image patch similarity (LPIPS). Higher PSNR and SSIM and lower LPIPS indicate greater agreement with the corresponding reference images.

\subsubsection{Near-field reconstruction fidelity across workflow conditions}

Because ground-truth images from arbitrary virtual viewpoints were unavailable, we used the spatially adjacent right-camera images as a standardized reference to compare reconstruction fidelity across workflow phases and recording sites. For each stereo pair, the right image contributed to the stereo-depth estimate used for scene initialization but was withheld from photometric supervision during 3DGS optimization, which used only the left-camera images. The resulting representation was then rendered from each calibrated right-camera pose, separated from the corresponding left camera by a baseline of approximately 12~cm. We computed PSNR, SSIM, and LPIPS between the rendered and captured right-camera images to characterize near-field reconstruction fidelity across workflow conditions.

\subsubsection{Egocentric fidelity against paired PoV recordings}

To quantitatively compare the egocentric view quality, a staff member carried a hand-held PoV camera during routine OR activities at Site A, with the wall-mounted camera system recording concurrently. For each of the ten evaluated scenes, the PoV camera pose was manually specified by aligning the virtual camera with the approximate physical location and orientation of the hand-held camera. EgoSurg then rendered an image using the registered pose and intrinsics matched to the PoV camera. The synthesized views were compared directly with the corresponding PoV recordings using PSNR, SSIM, and LPIPS. 

\subsubsection{Temporal stability}

To characterize temporal stability, we reconstructed sequences of 10 consecutive timestamps from each workflow phase and rendered them from the fixed, calibrated right-camera viewpoint. Temporal PSNR (t-PSNR), temporal SSIM (t-SSIM), temporal LPIPS (t-LPIPS), and flicker score were computed between consecutive rendered frames and averaged across the nine adjacent frame pairs in each sequence. Higher t-PSNR and t-SSIM and lower t-LPIPS indicate greater image-space continuity between adjacent timestamps. The flicker score characterizes frame-to-frame variations in luminance and local image structure, with lower values indicating fewer visible flickering artifacts. 

\section{Results}

\subsection{Deployment Across OR Workflows}

\begin{figure}[t]
  \centering
  \includegraphics[width=\textwidth]{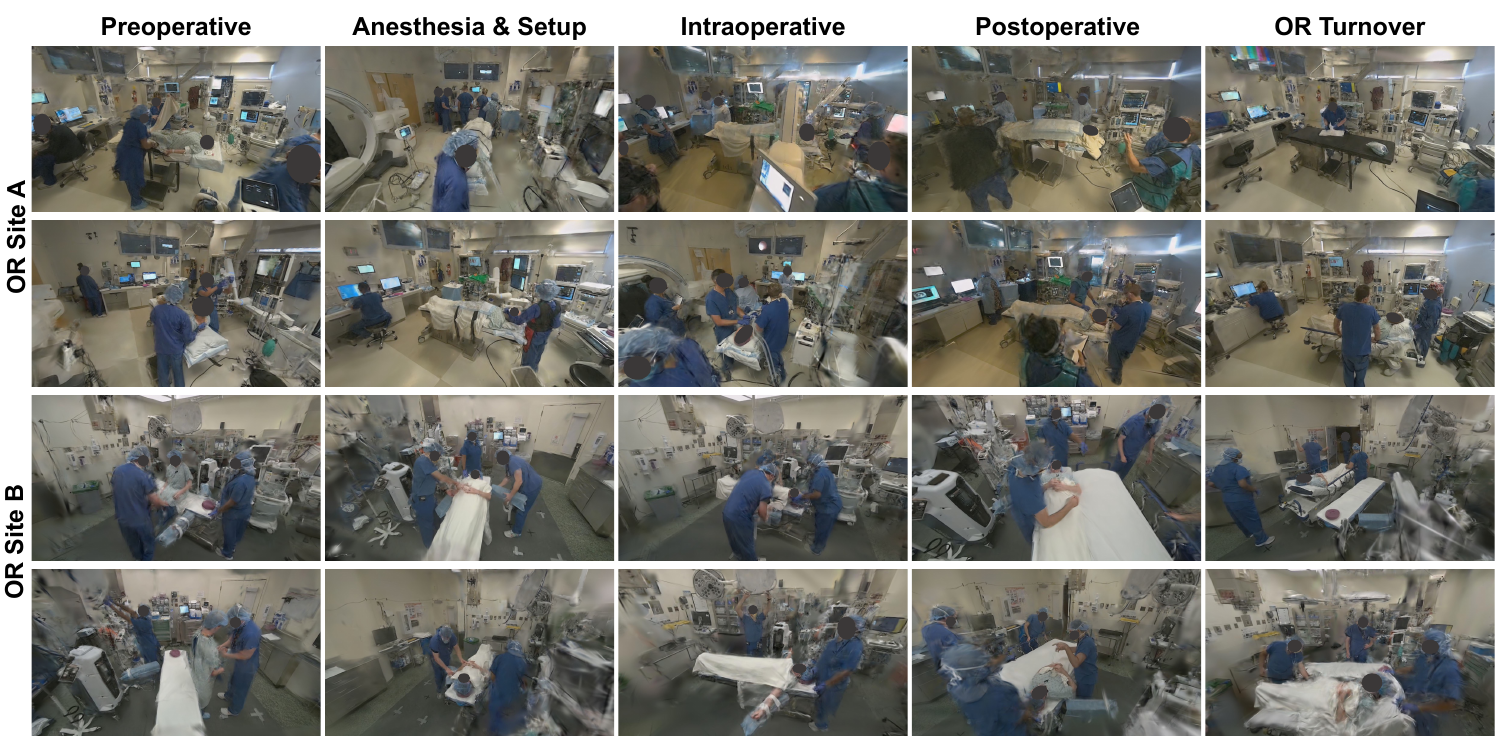}
  \caption{Qualitative results of arbitrary view rendering across different phases and OR sites.}  
  \label{arbitrary_result}
\end{figure}

EgoSurg was successfully deployed during four full-length robotic pulmonology procedures at Site A, with continuous recording from patient entry through room turnover, and during two simulated full-workflow sessions at Site B. Capture required no instrumentation of personnel and no equipment within the sterile field. System installation and calibration required approximately \SI{30}{min} at each site, were performed once, and required no adjustment between subsequent recordings. Reconstruction succeeded for all 150 sampled scenes (Figure~\ref{arbitrary_result}). Reconstruction was performed offline in approximately \SI{2}{min} per timestamp, after which novel views could be rendered at more than \SI{150}{fps}.

Near-field reconstruction fidelity remained comparable across workflow phases and recording sites (Table~\ref{tab:novelview_phases_sites}). Across the five phases evaluated at Site A, PSNR varied by 0.8~dB, from 26.4~dB during the intraoperative phase to 27.2~dB during OR turnover. The intraoperative phase also produced the lowest SSIM (.883) and highest LPIPS (.159), consistent with the increased crowding, motion, and occlusion present during this phase. Performance at Site B was comparable to that at Site A (PSNR 26.6 versus 26.8~dB, SSIM .906 versus .895, LPIPS .149 versus .143). These findings demonstrate transfer across differences in room geometry and equipment configuration without site-specific retuning. Since Site B consisted of simulated workflows with lower personnel density, the cross-site results primarily demonstrate robustness to differences in recording environments.

\begin{table}[t]
\centering
\small
\setlength{\tabcolsep}{6pt}
\caption{Near-field reconstruction fidelity across workflow phases and recording sites. Values are reported as mean $\pm$ standard deviation.}
\label{tab:novelview_phases_sites}
\vspace{4pt}

\begin{tabular}{lccc}
\toprule
\textbf{Condition}
& \textbf{PSNR$\uparrow$}
& \textbf{SSIM$\uparrow$}
& \textbf{LPIPS$\downarrow$} \\
\midrule

\multicolumn{4}{l}{\textit{Workflow phase}} \\
Preoperative
& 26.9 $\pm$ 0.2 & .898 $\pm$ .003 & .139 $\pm$ .004 \\
Anesthesia \& setup
& 26.8 $\pm$ 0.2 & .894 $\pm$ .003 & .143 $\pm$ .005 \\
Intraoperative
& 26.4 $\pm$ 0.4 & .883 $\pm$ .007 & .159 $\pm$ .014 \\
Postoperative
& 26.8 $\pm$ 0.4 & .896 $\pm$ .008 & .142 $\pm$ .013 \\
OR turnover
& 27.2 $\pm$ 0.2 & .903 $\pm$ .005 & .134 $\pm$ .005 \\
\midrule

\multicolumn{4}{l}{\textit{Recording site}} \\
Site A
& 26.8 $\pm$ 0.4 & .895 $\pm$ .009 & .143 $\pm$ .013 \\
Site B
& 26.6 $\pm$ 0.4 & .906 $\pm$ .003 & .149 $\pm$ .004 \\

\bottomrule
\end{tabular}
\end{table}

EgoSurg outperformed all baseline and ablated variants in near-field reconstruction fidelity (Table~\ref{tab:novelview_baseline}). Compared with image-only 3DGS, depth-based initialization increased near-field PSNR from 18.0 to 24.3~dB and SSIM from .674 to .857, while reducing LPIPS from .416 to .182. Adding diffusion-refined auxiliary views further increased PSNR to 26.8~dB and SSIM to .895 and reduced LPIPS to .143. EgoSurg also achieved the strongest frame-to-frame continuity, reaching a t-PSNR of 31.7~dB and a t-SSIM of .941. Overall, these results show that stereo-depth initialization primarily improves spatial reconstruction fidelity, while both depth initialization and diffusion-refined supervision contribute to smoother image-space transitions across independently reconstructed timestamps.

\begin{table}[t]
\centering
\small
\setlength{\tabcolsep}{6pt}
\caption{Comparison with baseline methods and the ablated variant in near-field reconstruction fidelity and temporal stability. Values are reported as mean $\pm$ standard deviation.}
\label{tab:novelview_baseline}
\vspace{4pt}
\begin{tabular}{lccc}
\multicolumn{4}{c}{\textbf{(A) Near-field reconstruction fidelity}} \\[1pt]
\toprule
\textbf{Method}
& \textbf{PSNR$\uparrow$}
& \textbf{SSIM$\uparrow$}
& \textbf{LPIPS$\downarrow$} \\
\midrule
Depth reprojection
& 16.4 $\pm$ 0.2 & .379 $\pm$ .014 & .800 $\pm$ .010 \\
Image-only 3DGS
& 18.0 $\pm$ 0.3 & .674 $\pm$ .013 & .416 $\pm$ .014 \\
EgoSurg w/o diffusion
& 24.3 $\pm$ 0.4 & .857 $\pm$ .011 & .182 $\pm$ .013 \\
\textbf{EgoSurg}
& \textbf{26.8 $\pm$ 0.4}
& \textbf{.895 $\pm$ .009}
& \textbf{.143 $\pm$ .013} \\
\bottomrule
\end{tabular}

\vspace{8pt}
\begin{tabular}{lcccc}
\multicolumn{5}{c}{\textbf{(B) Temporal stability}} \\[1pt]
\toprule
\textbf{Method}
& \textbf{t-PSNR$\uparrow$}
& \textbf{t-SSIM$\uparrow$}
& \textbf{t-LPIPS$\downarrow$}
& \textbf{Flicker$\downarrow$} \\
\midrule
Depth reprojection
& 16.8 $\pm$ 0.8 & .341 $\pm$ .031 & .272 $\pm$ .008 & .073 $\pm$ .009 \\
Image-only 3DGS
& 27.9 $\pm$ 0.9 & .883 $\pm$ .020 & .134 $\pm$ .009 & .022 $\pm$ .003 \\
EgoSurg w/o diffusion
& 28.7 $\pm$ 0.7 & .897 $\pm$ .014 & .085 $\pm$ .009 & .018 $\pm$ .002 \\
\textbf{EgoSurg}
& \textbf{31.7 $\pm$ 0.7}
& \textbf{.941 $\pm$ .009}
& \textbf{.050 $\pm$ .005}
& \textbf{.013 $\pm$ .001} \\
\bottomrule
\end{tabular}

\end{table}

\subsection{Egocentric Fidelity Against Ground Truth}
\label{sec:ego}

\begin{figure}[t]
\centering
\begin{minipage}{\textwidth}
\centering
\small

\captionof{table}{Egocentric rendering compared against paired ground-truth recordings
from a hand-held camera.}
\label{tab:egocentric_quant}
\vspace{4pt}
\begin{tabular}{lccc}
\toprule
\textbf{Method} & \textbf{PSNR$\uparrow$} & \textbf{SSIM$\uparrow$} & \textbf{LPIPS$\downarrow$} \\
\midrule
Depth reprojection & 13.7 $\pm$ 1.8 & .239 $\pm$ .059 & .933 $\pm$ .032 \\
Image-only 3DGS & 15.1 $\pm$ 1.2 & .694 $\pm$ .033 & .630 $\pm$ .037 \\
EgoSurg w/o diffusion & 16.2 $\pm$ 2.0 & .641 $\pm$ .057 & .520 $\pm$ .054 \\
\textbf{EgoSurg} & \textbf{17.8 $\pm$ 2.0} & \textbf{.766 $\pm$ .026} & \textbf{.406 $\pm$ .062} \\
\bottomrule
\end{tabular}
\end{minipage}
\vspace{1em}

\includegraphics[width=\textwidth]{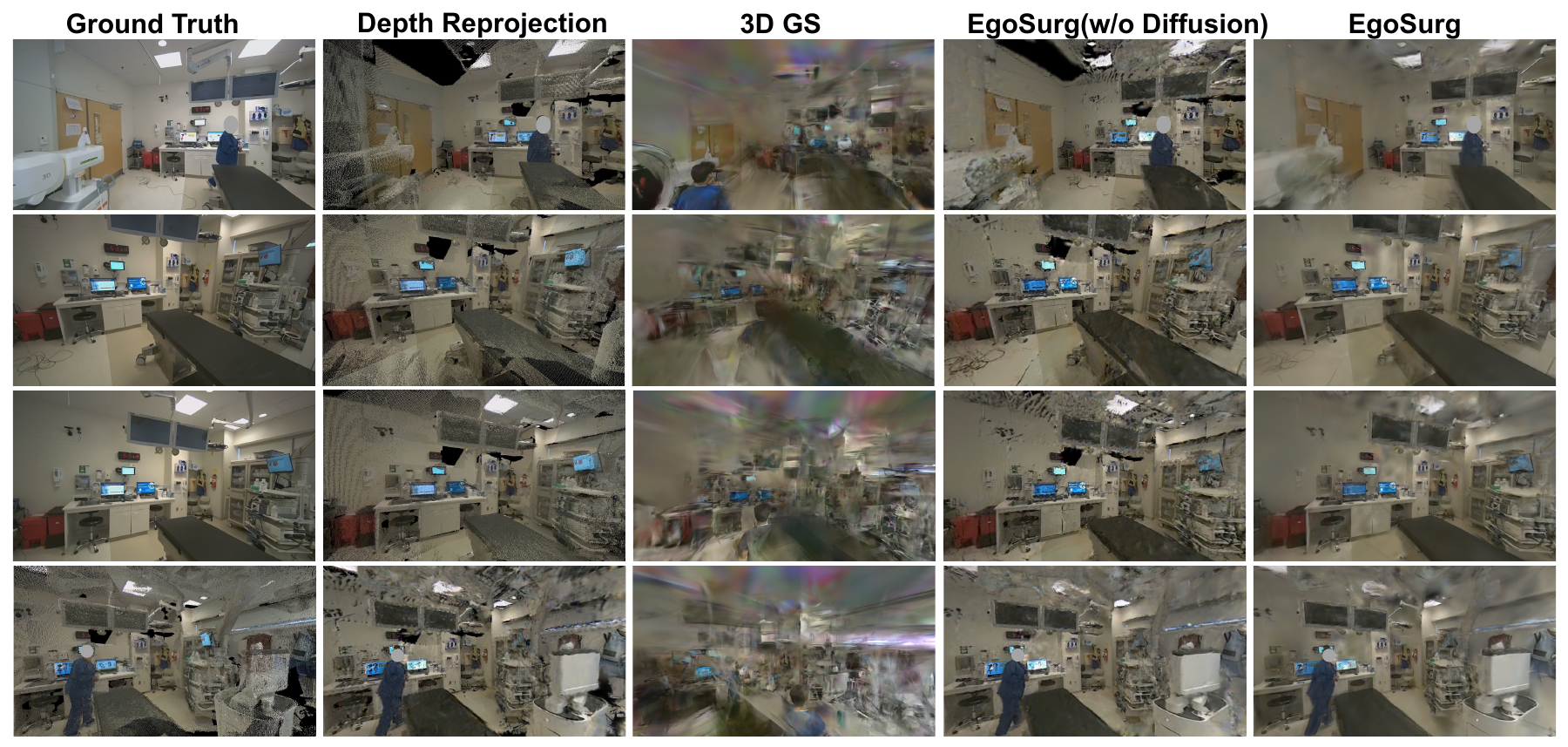}
\caption{Qualitative comparison of egocentric view rendering across methods, with hand-held camera recordings as ground truth.}
\label{pov_result}
\end{figure}

Against paired hand-held PoV recordings, EgoSurg achieved a PSNR of 17.8~dB, an SSIM of .766, and an LPIPS of .406, outperforming all comparison methods (Table~\ref{tab:egocentric_quant}). Relative to EgoSurg without diffusion, the complete method improved PSNR by 1.6~dB, increased SSIM from .641 to .766, and reduced LPIPS from .520 to .406. Performance was lower under the egocentric protocol than under the near-field protocol because the target PoV camera was farther from the ambient cameras and often closer to scene content, resulting in larger viewpoint changes and more pronounced occlusion and disocclusion. In addition, the reported pixel-level metrics reflect both reconstruction error and residual pose-registration error from manual camera alignment. Qualitatively, the synthesized views preserved the overall room layout, personnel locations, and equipment configuration more effectively than the comparison methods (Figure~\ref{pov_result}). Diffusion-based refinement reduced black gaps and incomplete-region artifacts in the unrefined renderings, as well as the floaters observed with image-only 3DGS. However, residual artifacts remained near scene boundaries and in regions with limited ambient-camera coverage.

\subsection{Case Studies}

\subsubsection{Retrospective adjudication of a simulated sterile field violation}

\begin{figure}[t]
  \centering
  \includegraphics[width=0.8\textwidth]{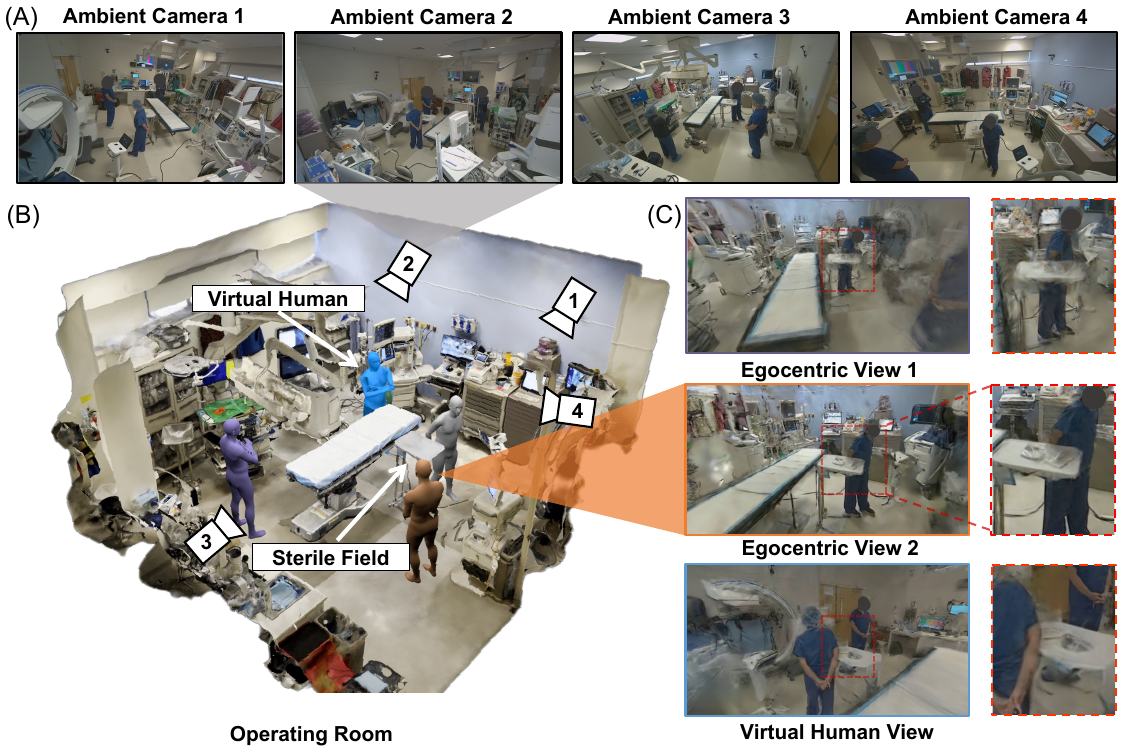}
  \caption{Retrospective adjudication of a simulated sterile field violation. \textbf{(A)} Four ambient wall-mounted cameras provide complementary but partially occluded views of the OR. \textbf{(B)} EgoSurg reconstructs the scene with a simulated sterile field, in which an arm contacts the sterile field boundary; a virtual observer can then be positioned at an unoccluded vantage point. \textbf{(C)} Egocentric replays from different team members' approximate viewpoints reveal the contact that is occluded or ambiguous in the original camera feeds.}  
  \label{app_safety}
\end{figure}
 
Ensuring safety in the OR is paramount, yet many near misses and critical events, such as sterile field violations, remain undocumented or rely on reports, limiting systematic review and prevention. Traditional video systems fall short when they do not capture the vantage points needed to determine whether safety boundaries were breached (Figure~\ref{app_safety}A). EgoSurg addresses this challenge by reconstructing the immersive OR scene and enabling retrospective replay from virtual viewpoints (Figure~\ref{app_safety}B). These additional viewpoints reveal spatial relationships that may remain occluded or ambiguous in the original camera feeds. In the simulated sterile field violation, EgoSurg exposed the contact between the participant's arm and the sterile field boundary from multiple viewpoints (Figure~\ref{app_safety}C), providing visual documentation that could support retrospective adjudication. This capability may allow otherwise anecdotal safety events to be examined more systematically during quality review.

\subsubsection{Role-specific replay of a robotic pulmonology case}

\begin{figure}[t]
  \centering
  \includegraphics[width=0.8\textwidth]{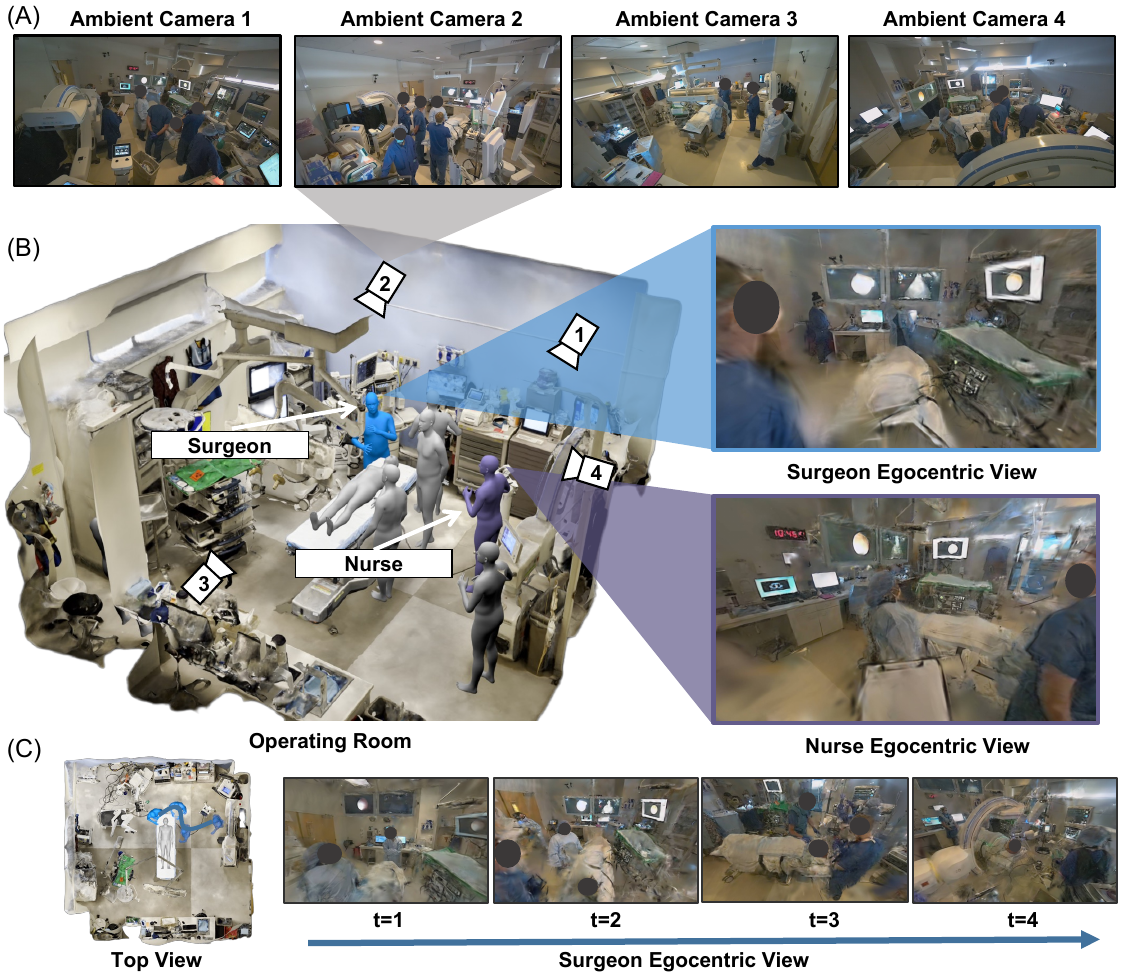}
  \caption{Role-specific egocentric replay of a robotic pulmonology case. \textbf{(A)} Four ambient wall-mounted cameras capture the procedure during the endobronchial ultrasound phase. \textbf{(B)} The reconstructed scene shows the team gathered around a monitor, while virtual egocentric views reflect what was visible to each role at that moment. \textbf{(C)} Replay of the surgeon's trajectory and corresponding egocentric views.}
  \label{app_training}
\end{figure}

Surgical education is fundamentally experiential. However, operative notes, fixed-camera recordings, and retrospective debriefs do not capture how each team member’s physical position shaped their visual access to instruments, displays, and other personnel. We examined a robotic pulmonology case during the endobronchial ultrasound phase, in which the team gathered around a shared monitor (Figure~\ref{app_training}A). Virtual cameras were manually positioned at the approximate head locations of team members, allowing EgoSurg to reconstruct the visual information available from different role-specific viewpoints. Replaying a specified trajectory approximating the surgeon's movement across the phase yielded a sequence of egocentric views showing how access to instruments and monitors changed with position (Figure~\ref{app_training}C). Unlike wearable cameras, which must be assigned prospectively and capture only the selected person's recorded trajectory, EgoSurg generated multiple role-specific views retrospectively from a single ambient recording. Such perspective-shifting replays could support both role-specific and cross-role training by allowing learners to examine a workflow from the physical vantage points of different team members.

\subsubsection{Counterfactual repositioning for line-of-sight analysis}

\begin{figure}[t]
  \centering
  \includegraphics[width=0.8\textwidth]{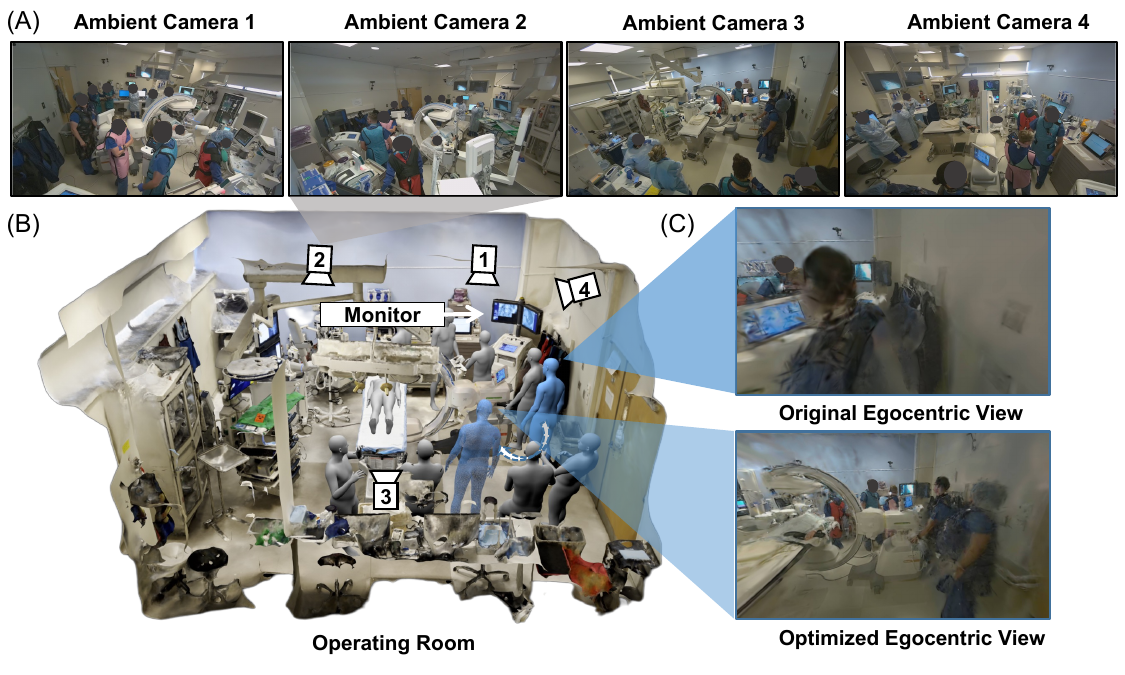}
  \caption{Counterfactual repositioning for line-of-sight analysis. \textbf{(A)} Four ambient wall-mounted cameras capture a robotic pulmonology case under C-arm fluoroscopy. \textbf{(B)} The reconstructed OR shows multiple team members oriented toward the X-ray monitor. One participant's line of sight (blue) is blocked, whereas a counterfactual position \SI{1}{m} left-posterior (light blue) leaves the monitor unoccluded without obstructing others. \textbf{(C)} Egocentric replays comparing the original and repositioned viewpoints.}  
  \label{app_efficiency}
\end{figure}

Limited visual access to shared information sources, such as fluoroscopy monitors during robotic pulmonology, can create bottlenecks in crowded OR environments. Conventional fixed-camera footage shows only the recorded viewpoints and cannot reveal whether an alternative position would have improved access to information. In a biopsy under C-arm fluoroscopy, the X-ray monitor was the focal information source, but one team member's sightline was intermittently blocked by other personnel (Figure~\ref{app_efficiency}A.B). We anchored a virtual camera at the team member's approximate location and generated a counterfactual viewpoint \SI{1}{m} left-posterior to the original position, while holding the reconstructed scene fixed. Re-rendering the scene from the two viewpoints showed that the monitor was visible from the counterfactual position during intervals in which it was occluded from the original position (Figure~\ref{app_efficiency}C). Because both views were generated from the same reconstruction, EgoSurg enabled the alternative viewpoint to be evaluated without re-staging the case. By making visibility and occlusion directly inspectable, EgoSurg turns a qualitative hunch ``move left and back'' into concrete visual evidence for evaluating layout and stance adjustments that may improve team coordination.

\section{Discussion}

This work demonstrates that sparse ambient cameras can reconstruct OR workflows as navigable 3D scenes, enabling retrospective observation from viewpoints that were not directly recorded. EgoSurg was successfully deployed across multiple workflows in two hospital ORs and maintained comparable near-field reconstruction fidelity across workflow phases and recording environments. Egocentric view fidelity was further evaluated against paired hand-held PoV recordings, yielding a PSNR of 17.8~dB, an SSIM of .766, and an LPIPS of .406. Beyond quantitative reconstruction performance, the case studies demonstrate several practical uses of EgoSurg. In the simulated sterile field violation, a novel viewpoint revealed a contact event that was difficult to assess from the fixed-camera recordings. In the robotic pulmonology case, replaying the procedure from the viewpoints of different team members provided material for immersive and cross-role training. The reconstructed scene also allowed us to test alternative observer positions and assess how changes in personnel placement could affect visibility. These case studies show how post hoc viewpoint selection can extend the value of fixed-camera recordings beyond conventional video review.

The current framework has several limitations. First, each timestamp is reconstructed independently without an explicit spatiotemporal prior. Although EgoSurg improved the measured frame-to-frame continuity, small changes in geometry and appearance can remain across adjacent timestamps, particularly around moving personnel, equipment boundaries, and reflective surfaces. Independent reconstruction also makes continuous replay I/O-bound because a separate 3DGS representation must be loaded for each timestamp. Second, rendering fidelity decreases for viewpoints far from the ambient cameras or close to scene content, where limited parallax and extensive occlusion leave portions of the scene weakly observed. Diffusion refinement reduces missing-region artifacts but may introduce implausible appearance that is not directly supported by the captured images; reconstructed views should therefore be interpreted together with the original recordings when reviewing fine spatial details or safety-critical events. Third, because role-specific viewpoints were manually specified from the approximate locations and orientations of team members, EgoSurg does not recover gaze, visual attention, or an individual's exact field of view. 

Future work will focus on spatiotemporal scene representations that support continuous replay, automatic estimation of personnel location and head orientation, and uncertainty measures that distinguish directly observed content from generative completion. Incorporating audio, gaze, and other workflow signals could further connect reconstructed viewpoints with the information and decisions available to each role. With these advances, ambient OR recordings could evolve from fixed archives into interactive records that allow surgical events to be revisited, compared, and analyzed from the viewpoints most relevant to safety, training, and workflow improvement.

\section*{Data Availability Statement}
The video recordings analyzed in this study contain identifiable patient and operating room personnel information and cannot be made publicly available because of privacy restrictions. De-identified data supporting the conclusions of this article and code used for reconstruction and evaluation are available from the corresponding author upon reasonable request.

\section*{Ethics Statement}
All data collection procedures in this study were reviewed and approved by the appropriate Institutional Review Board (IRB00421946 and HIRB00016983). Written informed consent was obtained from all patients whose procedures were recorded and from all operating room personnel who appeared in the recordings.

\section*{Author Contributions}

H.Z. and M.U. contributed to the main ideas and methods. H.Z., L.S., H.D. and H.S. contributed to the experimental design. H.Z., L.S., J.L.P., A.G., B.D.K., J.K.J. and A.C.A. contributed to data collection. M.U., J.L.P., A.C.A., J.K.J., M.I. and L.Y. contributed to resources, funding, hardware and facilities. H.Z., L.S., M.U. and R.D.S. drafted the manuscript.

\section*{Funding}
This work was supported by the National Science Foundation under Award No.~2239077 and by Johns Hopkins University Internal Funds.

\section*{Conflict of Interest Statement}
The authors declare that the research was conducted in the absence of any commercial or financial relationships that could be construed as a potential conflict of interest.

\bibliographystyle{Frontiers-Harvard} 
\bibliography{bibliography}
\end{document}